\documentclass{article}

\usepackage{microtype}
\usepackage{graphicx}
\usepackage{subfigure}
\usepackage{booktabs} 
\usepackage[table,dvipsnames]{xcolor}
\usepackage{colortbl}
\usepackage{enumerate}

\usepackage{hyperref}

\usepackage[accepted]{icml2025}

\usepackage{amsmath}
\usepackage{amssymb}
\usepackage{mathtools}
\usepackage{amsthm}

\usepackage[capitalize,noabbrev]{cleveref}

\theoremstyle{plain}

\theoremstyle{definition}

\theoremstyle{remark}

\usepackage[textsize=tiny]{todonotes}

\begin{document}

\twocolumn[
\icmltitle{BlockBPE: Parallel BPE Tokenization}



\icmlsetsymbol{equal}{*}

\begin{icmlauthorlist}
\icmlauthor{Amos You}{berkeley}
\end{icmlauthorlist}

\icmlaffiliation{berkeley}{UC Berkeley}

\icmlcorrespondingauthor{Amos You}{amosyou@berkeley.edu}

\icmlkeywords{Machine Learning, ICML}

\vskip 0.3in
]



\printAffiliationsAndNotice{}  

\begin{abstract}
    Tokenization is a critical preprocessing step in large language model pipelines, yet widely-used implementations remain CPU-bound and suboptimal for batch inference workflows on GPU. We present BlockBPE, a parallel GPU implementation of byte-pair encoding (BPE) that achieves near linear-time complexity under realistic assumptions and is optimized for high-throughput, batch inference. Unlike existing Rust-based tokenizers such as HuggingFace Tokenizers or OpenAI's tiktoken—whose runtimes are dominated by Regex pre-tokenization and exhibit $O(n \log n)$ runtime—BlockBPE eliminates the Regex pre-tokenization which leads to small loss in generation quality, but enables highly parallelized token merges within thread blocks, reducing overall complexity to $O(nd)$ where $d \ll n$. On high-batch inference workloads, BlockBPE achieves up to 2x higher throughput than tiktoken and 2.5x over HuggingFace Tokenizers.
\end{abstract}

\section{Introduction}


Tokenization is a foundational step in natural language processing pipelines, responsible for transforming raw text into discrete tokens that serve as the inputs to language models. In large language models, byte-pair encoding (BPE) tokenization is standard due to its abililty to handle rare or unknown words while keeping the overall vocabulary size manageable. Despite its wide usage, BPE implementations remain largely CPU-bound, creating a performance bottleneck when processing millions of tokens on heterogenous hardware.

Existing BPE tokenizers rely on Regex pre-tokenization to split text into coarse units of words, numbers, and punctuation, before fine-grain byte-level merges happen using learned rules. This sequential, branching logic is well suited for CPUs, but limits scalability on massively parallel hardware. In constrast, modern AI systems have scaled up to leverage GPU compute in order to meet the demand of larger models and datasets. As model sizes and dataset sizes continue to grow, many steps in the LLM inference pipeline—including tokenization—become performance critical and appropriate for parallelization on GPUs. 

In spite of growing interest in end-to-end GPU acceleration for deep learning workloads, tokenization has largely been left out of this discussion. Open-source LLM serving engines such as vLLM \cite{kwon2023efficient} and SGLang \cite{zheng2024sglang} have focused their efforts on online inference settings, where low latency on small batch sizes is crucial—yet omitting batch tokenization introduces a missed opportunity for higher throughput. vLLM still lacks CPU-based batch tokenization, whereas SGLang has only recently added it to boost end-to-end throughput. By moving tokenization onto the GPU, we can further streamline the inference pipeline and unlock new opportunities for co-optimizing tokenization alongside model execution.


In this work,

\vspace{-0.25cm}
\begin{enumerate}[(1)]
    \item we propose BlockBPE, the first GPU implementation of BPE tokenization, using GPU-resident hashmaps and block-level prefix scans
    \vspace{-0.25cm}
    \item we examine the settings in which a GPU implementation outperforms CPU-based implementations
    \vspace{-0.25cm}
    \item we identify the limitations of BlockBPE as well as the algorithmic and hardware constraints that remain.
\end{enumerate}



\section{Related Work}
\label{related_work}

\subsection{Tokenization Methods}

Early research in Neural Machine Translation (NMT) laid the groundwork for the tokenizers we know today. Byte-Pair Encoding (BPE) was popularized by \cite{sennrich-etal-2016-neural}, which demonstrated that merging frequent byte-pair units enables efficient coverage of rare and out-of-vocabulary words. WordPiece \cite{wu2016googlesneuralmachinetranslation} has a similar vocabulary training process as BPE, but a greedy, longest-prefix match is used rather than sequentially merging adjacent tokens. SentencePiece \cite{kudo-richardson-2018-sentencepiece} uses expectation-maximization and filter pruning to learn a vocabulary of subwords with probabilities, and probablistically segments the input during tokenization. We focus on BPE tokenization specifically, as most decoder-only LLMs currently use BPE tokenizers.

\subsection{Libraries}

The most widely adopted tokenization libraries are HuggingFace Tokenizers \cite{Moi_HuggingFace_s_Tokenizers_2023} and OpenAI’s tiktoken \cite{tiktoken}. HuggingFace Tokenizers is the preferred choice for model training and inference due to its integration with HuggingFace Transformers and how weights can be hosted on HuggingFace Hub. tiktoken is favored by developers who need precise control over token counts when interacting with OpenAI’s API. By tokenizing inputs locally before making API calls, users can accurately estimate and limit the number of tokens they send.

In addition to tokenizer libraries, cuDF \cite{cudf} is a GPU DataFrame library that also has some tokenization functionality. cuDF has support for WordPiece tokenization, which allows for 483x faster BERT tokenization than HuggingFace's Rust implementation \cite{Jawa_2021}. However, cuDF does not have BPE fully implemented, with the only functionality being string transformations into tokens, with no token to ID lookup or processing to output a tensor. This functionality is more targeted towards offline data processing, such as tokenizing training data before large training runs.

\section{Background}
\label{background}

We provide some background on how the BPE algorithm works, as well as how it is implemented and optimized for CPU.

\subsection{Naive BPE}

Given an input string $S$, the BPE algorithm (Algorithm \ref{alg:bpe_tokenization_merge_table}) splits $S$ into a list $T$ of individual bytes or tokens, known as pre-tokenization. Through multiple merge passes, each pass will find the pair of adjacent tokens with the lowest rank using a learned merge table $M$ and merges the two into one token. This process can be done with $O(n^2)$ runtime. Each merge pass requires scanning $O(n)$ token pairs, lookups from a hashmap for merge rank of the token pair is $O(1)$ time, and applying the merges takes $O(n)$ time. In the worst case, we will have $O(n)$ merge passes, where all initial tokens end up merging down to a single token in the vocabulary. Overall, this is $O(n^2)$ runtime.

\begin{algorithm}[h]
    \caption{BPE Tokenization with Merge Table}
    \label{alg:bpe_tokenization_merge_table}
\begin{algorithmic}
    \REQUIRE Input string $S$, merge table $M$ with list of token pairs and rank
    \STATE Split $S$ into a list of bytes: $T \gets \text{list}(S)$
    \STATE Initialize a map $R$ from pair $(a, b)$ to its merge rank from $M$
    \WHILE{true}
        \STATE Build list of adjacent token pairs in $T$
        \STATE Find the pair $(a, b)$ in $T$ with the lowest rank in $R$
        \IF{no such pair exists}
            \STATE \textbf{break}
        \ENDIF
        \STATE Replace left-most occurrence of $(a, b)$ in $T$ with merged token $ab$
    \ENDWHILE
    \STATE \textbf{return} Final list of tokens $T$
\end{algorithmic}
\end{algorithm}

\subsection{CPU Implementations}

The naive BPE algorithm does byte-level pre-tokenization, but in practice all of the CPU implementations use Regex pre-tokenization. This allows string splitting to preserve certain patterns like \textbackslash{n}, 's, 't, long sequences of numbers, etc. Regex matching splits the string into manageable chunks, and rather than applying the sequential BPE merges on the full string, the merges occur within each Regex match. Moreover, CPU implemenations are able to achieve $O(n \log n)$ runtime by maintaining a priority queue of seen token pairs and corresponding ranks, and during each merge pass popping from the priority queue will give the next token pair to merge.

\section{BlockBPE}
\label{flash_bpe}

We show how the BPE algorithm can be implemented on GPU. 

\subsection{Pre-Tokenization}

Regex matching is the bottleneck in CPU implementations, taking up to 75\% of total tokenization runtime when benchmarked. Given that Regex involves backtracking and deep branching logic, it is inherently hard to parallelize. We substitute Regex pre-tokenization with byte-level pre-tokenization and special character lookups, which can be up to 3x faster. Special character lookups are necessary in order to identify beginning-of-sequence (BOS) and end-of-sequence (EOS) tokens, as well as the other special characters introduced during model pre-training. We implement byte-level pre-tokenization in Rust, which involves reading in bytes of sequences and hashmap lookups for the corresponding token ID.


\begin{figure}[h]
\vskip 0.2in
\begin{center}
    \centerline{\includegraphics[width=\columnwidth]{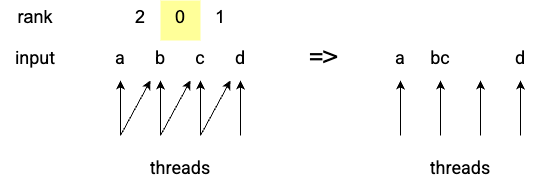}}
    \caption{\textbf{Merge pass from a thread perspective.} Each thread $i$ checks for the merge rank of token pair $(i, i+1)$. After a reduction across warps to find block minimum, corresponding threads apply the merge operation.}
\end{center}
\vskip -0.2in
\end{figure}

\subsection{Parallel Merge Passes}

Next, each merge pass can be highly parallelized. Suppose for an input string of length $n$, we have a block of $n$ threads that operate on each byte of the string. If we consider what work an individual thread $i$ has to do, it must

\vspace{-0.25cm}
\begin{enumerate}[(1)]
    \item read token $i$ and $i + 1$
    \vspace{-0.25cm}
    \item lookup map $M$ for rank of token pair $(i, i+1)$
    \vspace{-0.25cm}
    \item set minimum rank to the rank value if it is lower
\end{enumerate}
\vspace{-0.25cm}

This process can be parallelized with some careful thread synchronization to get the proper minimum value across the whole sequence. After the minimum rank is found, each thread $i$ must now

\vspace{-0.25cm}
\begin{enumerate}[(1)]
    \setcounter{enumi}{3}
    \item mark if it is at the token where merge should happen
    \vspace{-0.25cm}
    \item write token $i$ or the merged $(i, i+1)$
\end{enumerate}
\vspace{-0.25cm}

We can consider this as $O(1)$ time for each merge pass, giving us an overall $O(n)$ runtime if we have as many threads as the input string length. However, for long context settings, each thread will need to stride and move along the input string $d = (\text{seq\_len} / \text{block\_size})$ times, where block\_size = $n$. The overall runtime is $O(nd)$, and the ideal situation $d = 1$ occurs when our block can fully span the string.

\subsection{Implementation Details}

We utilize \texttt{cuCollections} for highly optimized concurrent hashmaps on GPU, and \texttt{CCCL} for optimized block-wide prefix scans to find the write indices for each thread in the block. Specifically, we keep a list of 0s/1s to mark the indices where the token has been merged (ie. token $i+1$). Then, we can do a exclusive prefix sum to do a compaction for the new write indices. In other words, if we are merging a pair, all the tokens to the right of the merged pair should move to the left by 1 index, which is done by the compaction.

We opted to design the GPU kernel to operate on a 1 block for 1 string basis. This choice stems from how it is easier to synchronize threads within each block, with the assumption that there is higher latency if we use cooperative groups to synchronize across blocks.

\section{Experiments}
\label{experiments}

While individual CPU threads benefit from low-latency execution due to superior cache locality, GPU threads unlock performance by scaling massively in parallel. With this latency-throughput tradeoff in mind, our goal is to elucidate at what point does tokenization on GPU become preferred and practical. As such, we benchmark BlockBPE against CPU implementations across different thread counts, batch sizes, and sequence lengths. All benchmarks are performed on the Intel Xeon Platinum 8470 and H100 SXM 80GB.

\subsection{Microbenchmarks}

We conduct microbenchmarks by tokenizing excerpts from \textit{War and Peace} by \cite{tolstoy1867} using the GPT-2 tokenizer, comparing results across GPU block sizes (ie. thread count), batch sizes, and sequence lengths.

\begin{figure}[h]
\vskip 0.2in
\begin{center}
\centerline{\includegraphics[width=\columnwidth]{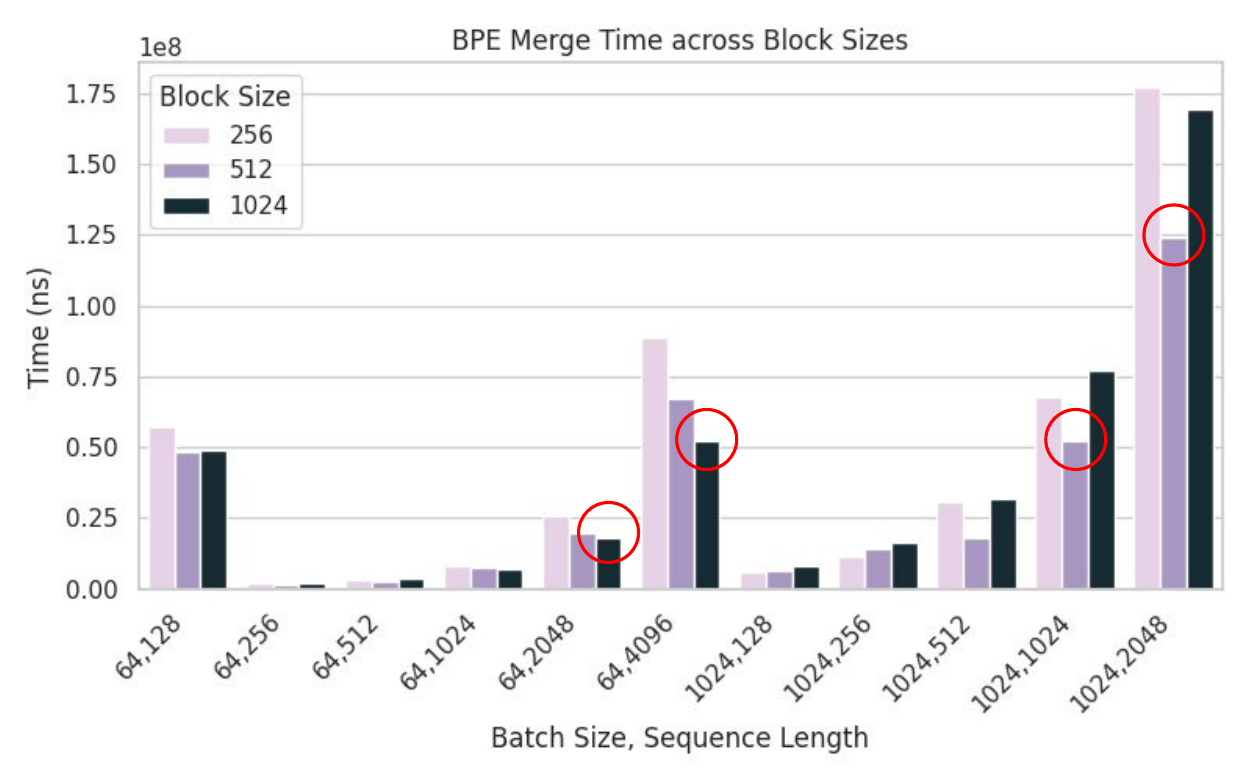}}
\caption{\textbf{BPE merge time comparison} for (256, 512, 1024) block sizes. Lower time is better.}
\label{block_size_merge_times}
\end{center}
\vskip -0.2in
\end{figure}

We find that there is a tradeoff between the batch size and sequence length which guides the optimal block size for BlockBPE (Figure \ref{block_size_merge_times}). For small batches of long sequences (64 x 2048/4096), a larger block size (1024) leads to faster merges. In the opposite setting where we have large batches of short sequences (1024 x 128/256), a smaller block size (256) leads to faster merges. This is expected since if we have long sequences, we want more threads to operate on the sequence in parallel. However, increasing the block size will decrease the number of thread blocks that can be executing simultaneously, since the number of resident thread blocks is fixed.



\begin{figure}[h]
\vskip 0.2in
\begin{center}
\centerline{\includegraphics[width=\columnwidth]{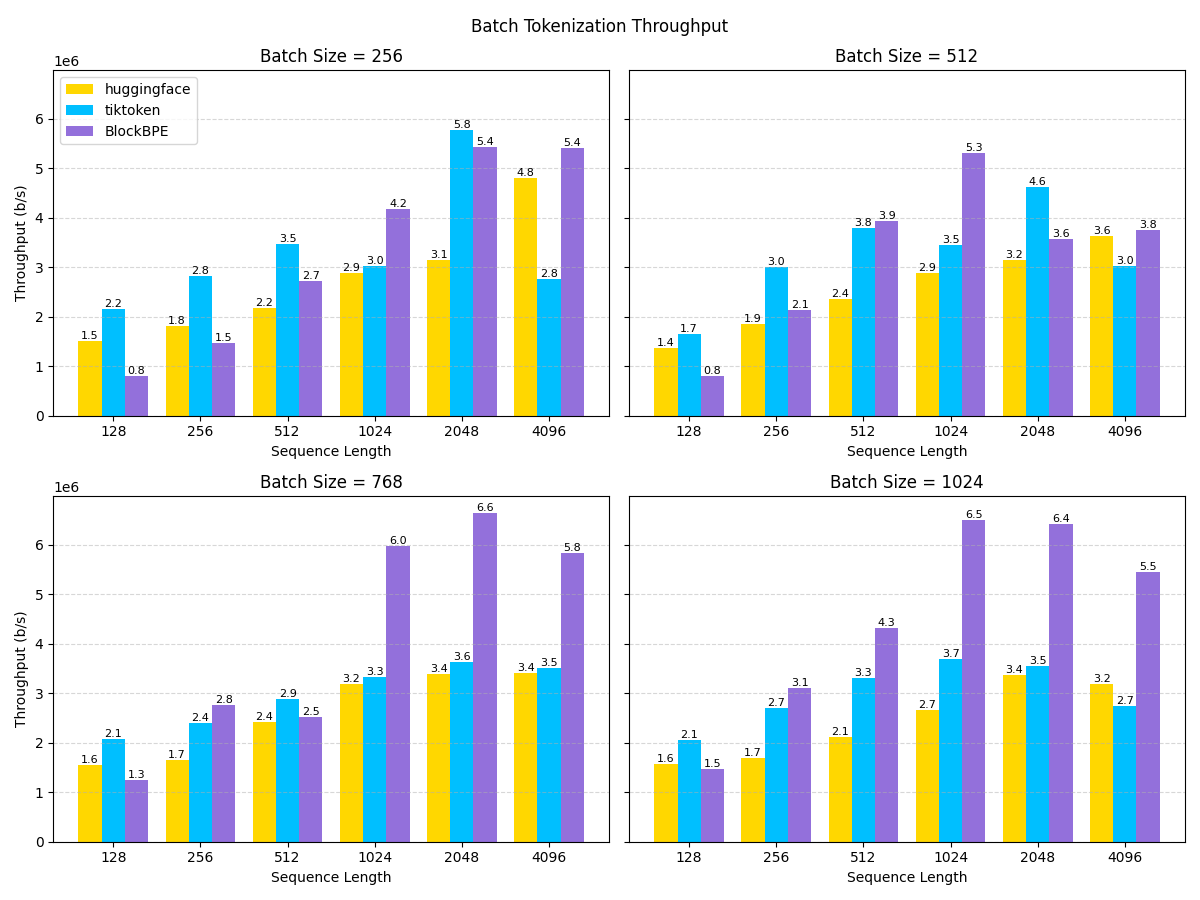}}
\caption{\textbf{Throughput comparison} between HuggingFace Tokenizers, tiktoken, and BlockBPE. BlockBPE achieves higher throughput in high batch, long sequence settings..}
\label{throughput_figure}
\end{center}
\vskip -0.2in
\end{figure}

BlockBPE achieves higher throughput than tiktoken and HuggingFace Tokenizers in all high batch size situations, ranging from batch sizes of 256-1024 (Figure \ref{throughput_figure}). With regards to the sequence length or the number of bytes in the input, BlockBPE's peak performance occurs when the block size is approximately the sequence length. This is ideal for BlockBPE because within each merge pass, all threads in a block can process the full string in parallel, without having to stride over the string multiple times. When the string is longer than the block size (eg. block size of 256, string length of 512), we employ thread coarsening where each thread will need to process multiple bytes from the input string. As we scale to extremely long sequences, each thread will have to do more work within each merge pass, leading to slower runtimes.

The number of inputs that each thread has to process will scale according to $d = (\text{seq\_len} / \text{block\_size})$. We can increase the block size to reduce the number of iterations $d$, but we do have a limit of how large our block size can be, since each block can have at most 1024 threads. Moreover, as we scale up block size, we have to reduce the batch size since each SM can schedule warps up to 1024 threads. On the H100 with a max block size of 1024 threads, our batch size can be 114 since we have 114 SMs. For block size of 256, we can scale to $114 \times (1024 / 256) = 528$ batch size.

\subsection{Tokenization Quality}

In many cases, byte-level pre-tokenization and Regex pre-tokenization will result in the same encodings, but there are some edge cases, such as repeated characters or 4-digit numbers, that are not encoded properly. For example, in the GPT-2 tokenizer, "...." should be one token but byte-level pre-tokenization will split the string to "." and "...". Or "1000" should be split as "100" "0" instead of "10" "00".

We can understand the tokenization quality of BlockBPE by comparing the similarity of the tokenized inputs with the ground truth from HuggingFace Tokenizers. For each input string $s_i$, we find the proportion of the string that matches between BlockBPE ($t_i^B$) and HF ($t_i^{HF}$) using the Levenshtein Distance ($d_L$), and average across all strings in the batch. Specifically, the similarity metric is as follows.

\[
\mathrm{sim}
\;=\;
\frac{1}{n}
\sum_{i=1}^{n}
1 - \frac{d_{L}\bigl(t_{i}^{\mathrm{HF}},\,t_{i}^{\mathrm{B}}\bigr)}{\lvert s_{i}\rvert}
\]

We compare the tokenized text and the evaluation accuracy between BlockBPE and HuggingFace Tokenizers on several datasets. MMLU \cite{hendryckstest2021} is a general multi-task language understanding dataset, GPQA \cite{rein2024gpqa} tests domain-expert level questions on biology, physics, and chemistry, GSM8K \cite{cobbe2021gsm8k} is a collection of grade school math word problems, and AGIEval \cite{zhong2023agieval} is compiled from various standardized exams across disciplines. These evaluations are done on Llama-3.1-8B-Instruct using lm-evaluation-harness \cite{eval-harness}.

\begin{table}[h]
\caption{\textbf{Similarity} of tokenized text and \textbf{Accuracy} between HF Tokenizers and BlockBPE (Llama-3.1-8B-Instruct)}
\label{similarity_table}

\vskip 0.15in
\begin{center}
\begin{small}
\begin{sc}
\newcommand{\lightgreen}{\rowcolor[green]{.90}}
\begin{tabular}{lccc}
    \toprule
    Dataset & Sim & Acc\textsubscript{HF} & Acc\textsubscript{BlockBPE} \\
    \midrule
    \rowcolor{green!10}
    MMLU & 0.999 & 0.692 $\pm$ 0.018 & 0.681 $\pm$ 0.018 \\
    \rowcolor{green!10}
    GPQA & 0.989 & 0.295 $\pm$ 0.022 & 0.295 $\pm$ 0.022 \\
    \rowcolor{red!10}
    GSM8K & 0.989 & 0.781 $\pm$ 0.012 & 0.224 $\pm$ 0.013 \\
    \rowcolor{green!10}
    AGIEval & 0.998 & 0.410 $\pm$ 0.044 & 0.410 $\pm$ 0.044 \\
    \bottomrule
\end{tabular}
\end{sc}
\end{small}
\end{center}
\vskip -0.1in
\end{table}



We discover that on general tasks and datasets comprised primarily of words like MMLU, GPQA, and AGIEval, BlockBPE maintains the same generation quality despite using byte-level pre-tokenization. However, for math evaluations like GSM8K, the mismatch in how numbers are tokenized leads to a noticeable performance drop by 56\%. 





\section{Future Directions and Conclusion}

\textbf{Future directions.} Future research on BlockBPE could explore enhancements to generation quality for LLM serving. By replacing Regex with byte-level pre-tokenization, we incur a loss in generation quality for downstream math tasks. Exploring alternative pre-tokenization schemes or using byte-level pre-tokenization during pre-training to mitigate this loss would be valuable. Integrations with LLM serving engines, including TensorRT-LLM, vLLM, and SGLang, would be beneficial to understand end-to-end performance, but requires major architecture modifications to support GPU tokenization. We leave this for future work.

\textbf{Conclusion.} We present BlockBPE, a parallel GPU implementation of byte-pair encoding (BPE) tokenizers. By replacing CPU-bound Regex pre-tokenization with byte-level pre-tokenization, we enable highly parallelized merge passes on GPU. We have shown that BlockBPE achieves higher throughput than existing tokenizer libraries HuggingFace Tokenizers and tiktoken on high batch inference setups.

\newpage

\section*{Acknowledgements}

I would like to thank Chuyi Shang and Dhruv Gautam for their insightful discussions and feedback. This work was supported in part by generous compute grants from Hyperbolic and Prime Intellect.

\section*{Impact Statement}

This paper presents work whose goal is to advance the field of 
Machine Learning. There are many potential societal consequences 
of our work, none which we feel must be specifically highlighted here.

\bibliography{main}
\bibliographystyle{icml2025}

\newpage



\end{document}